\documentclass[journal,twoside,web]{ieeecolor}
\usepackage{tmi}
\usepackage{cite}
\usepackage{amsmath,amssymb,amsfonts}
\usepackage{algorithmic}
\usepackage{graphicx}
\usepackage[warn]{textcomp}
\usepackage{hyperref}
\def\BibTeX{{\rm B\kern-.05em{\sc i\kern-.025em b}\kern-.08em
    T\kern-.1667em\lower.7ex\hbox{E}\kern-.125emX}}
\usepackage{xspace}

\def\eg{\textit{e.g.}~}

\def\etal{\textit{et al.}\xspace}
\def\ie{\textit{i.e.}~}

\def\etc{\emph{etc}.~} 
\def\vs{\emph{vs}.~}
 
\newcommand{\ra}[1]{\renewcommand{\arraystretch}{#1}}
\usepackage{color}

\newcommand{\JimSan}{Jim\'{e}nez-S\'{a}nchez}

\newcommand{\realnumbers}{\mathbb{R}}
\newcommand{\GitHubLink}{\href{https://github.com/ameliajimenez/curriculum-learning-prior-uncertainty}{https://github.com/ameliajimenez/curriculum-learning-prior-uncertainty}}

\usepackage{booktabs}% http://ctan.org/pkg/booktabs
\usepackage{tabu}
\usepackage{multirow}
\usepackage{subfig}
\usepackage{comment}

\DeclareMathOperator*{\argmin}{arg\,min}
\DeclareMathOperator*{\EX}{\mathbb{E}}% expected value

\begin{document}

\title{Curriculum learning for improved femur fracture classification: scheduling data with prior knowledge and uncertainty}

\author{Amelia \JimSan, Diana Mateus, Sonja Kirchhoff, Chlodwig Kirchhoff, \\ Peter Biberthaler, Nassir Navab, Miguel A. Gonz\'{a}lez Ballester, Gemma Piella
\thanks{This project has received funding from the European Union’s Horizon 2020 research and innovation programme under the Marie Sk\l{}odowska-Curie grant agreement No. 713673 and by the Spanish Ministry of Economy [MDM-2015-0502]. A. \JimSan{} has received financial support through the ``la Caixa'' Foundation (ID Q5850017D), fellowship code: LCF/BQ/IN17/11620013. D. Mateus has received funding from Nantes M\'etropole and the European Regional Development, Pays de la Loire, under the Connect Talent scheme. Authors thank Nvidia for the donation of a GPU.}
\thanks{A. JS., G. P., and M. A. G. B. are with BCN MedTech, Department of Information and Communication Technologies, Universitat Pompeu Fabra, 08018 Barcelona, Spain. (e-mails: amelia.jimenez@upf.edu, gemma.piella@upf.edu, ma.gonzalez@upf.edu). M. A. G. B. is also with ICREA, Barcelona, Spain.
}
\thanks{D. M. is with Ecole Centrale de Nantes, LS2N, UMR CNRS 6004, 44321, Nantes, France (e-mail: diana.mateus@ec-nantes.fr).}
\thanks{S. K., C. K., P. B. are with Department of Trauma Surgery, Klinikum rechts der Isar, Technische Universit{\"a}t M{\"u}nchen, 81675, Munich, Germany (e-mail: sonja.kirchhoff@me.com, dr.kirchhoff@me.com, peter.biberthaler@mri.tum.de). S. K. is also with Institute of Clinical Radiology, LMU M{\"u}nchen, Munich, Germany.}
\thanks{N. N. is with Computer Aided Medical Procedures, Technische Universit{\"a}t M{\"u}nchen, 85748, Munich, Germany (e-mail: nassir.navab@tum.de). N. N. is also with Johns Hopkins University,  21218, Baltimore, USA.}
}

\maketitle

\begin{abstract}
An adequate classification of proximal femur fractures from X-ray images is crucial for the treatment choice and the patients' clinical outcome. We rely on the commonly used AO system, which describes a hierarchical knowledge tree classifying the images into types and subtypes according to the fracture's location and complexity. In this paper, we propose a method for the automatic classification of proximal femur fractures into 3 and 7 AO classes based on a Convolutional Neural Network (CNN). As it is known, CNNs need large and representative datasets with reliable labels, which are hard to collect for the application at hand. In this paper, we design a curriculum learning (CL) approach that improves over the basic CNNs performance under such conditions. Our novel formulation reunites three curriculum strategies: individually weighting training samples, reordering the training set, and sampling subsets of data. The core of these strategies is a scoring function ranking the training samples. We define two novel scoring functions: one from domain-specific prior knowledge and an original self-paced uncertainty score. 
We perform experiments on a clinical dataset of proximal femur radiographs. The curriculum improves proximal femur fracture classification up to the performance of experienced trauma surgeons. The best curriculum method reorders the training set based on prior knowledge resulting into a classification improvement of 15\%. Using the publicly available MNIST dataset, we further discuss and demonstrate the benefits of our unified CL formulation for three controlled and challenging digit recognition scenarios: with limited amounts of data, under class-imbalance, and in the presence of label noise. The code of our work is available at: \scalebox{0.9}{\GitHubLink}.
\end{abstract}


\section{Introduction}
\label{sec:introduction}
\IEEEPARstart{P}{roximal} femur fractures are a significant cause of morbidity and mortality, giving rise to a notable socioeconomic impact \cite{Ryan2015, Giannoulis2016}. Elderly population in the western world are especially affected. The incidence of femur fractures increases exponentially from an age of 65 and is almost doubled every five years. 

%%%%%%%%%%%%%%%%%%%%%%%%%%%%%%%%%%%%%%%%%%%%%%%%%%%%% Figure 1:
\begin{figure*}[t]
    \centering
    \includegraphics[width=0.9\textwidth]{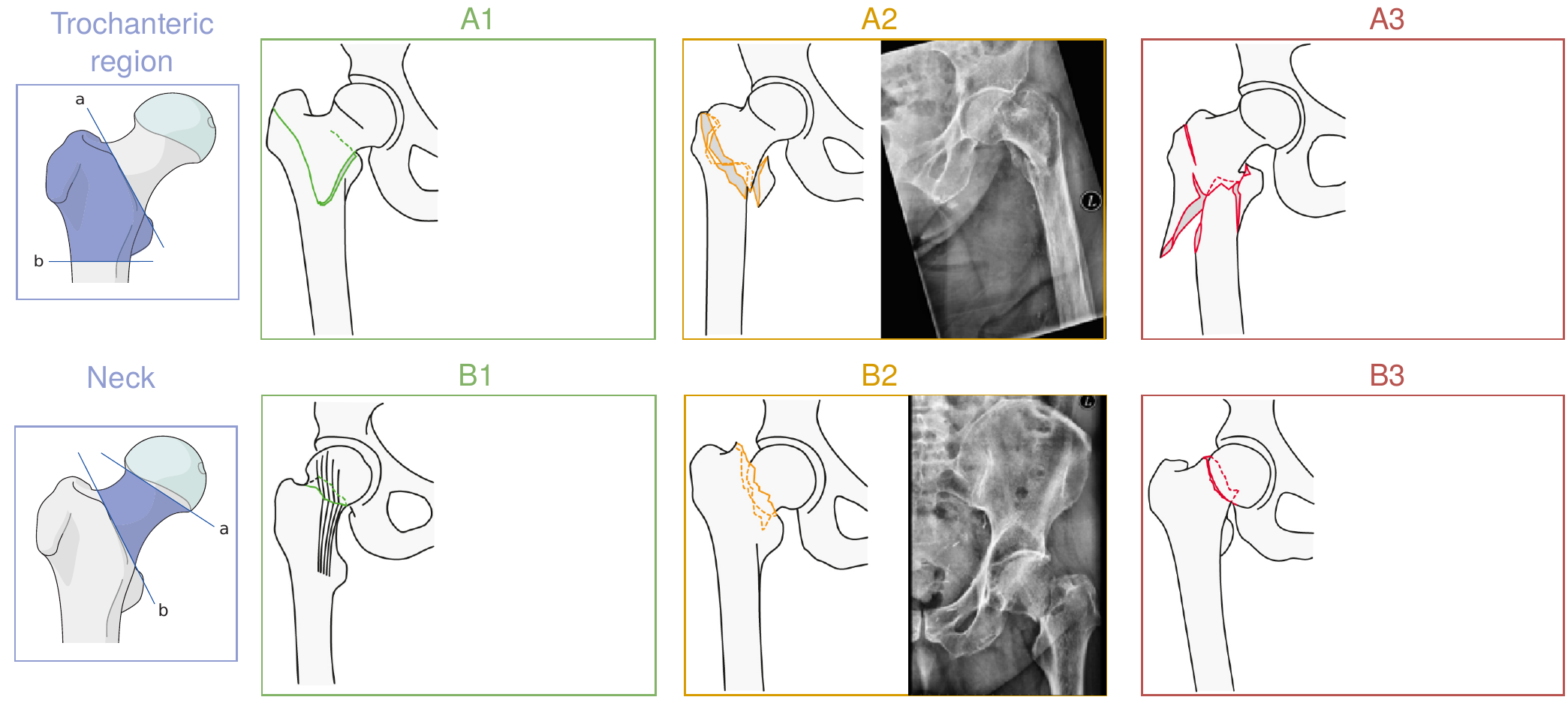}%
    \caption{Examples of proximal femur fractures and their fine-grained AO classification, adapted from~\cite{Kellam2018}.}%
    \label{fig:ao}%
\end{figure*}
%%%%%%%%%%%%%%%%%%%%%%%%%%%%%%%%%%%%%%%%%%%%%%%%%%%%%%%%%%%%%%%%%%%%%%

Surgery is the most common and preferred treatment for proximal femur fractures \cite{Sheehan2015}. The exact classification of the fracture is crucial for deciding the surgical procedure and choosing the surgical implant if needed. The Arbeitsgemeinschaft f{\"u}r Osteosynthesefragen (AO-Foundation) has established a hierarchical classification system for fractures of all bones based on radiographs. For proximal femur fractures, the AO classification has been beneficial, in terms of reproducibility, when compared against other systems such as the Jensen classification \cite{Bhandari2015}. The AO standard follows a hierarchy according to the location and configuration of the fracture lines, see Fig.~\ref{fig:ao}. Fractures of type-A are located in the trochanteric region, and fractures of type-B are those affecting the area around the femur neck. Each type of fracture is further divided into 3 subclasses depending on the morphology and number of fragments of the fracture.

The ability to adequately classify fractures according to the AO standard based on radiographs is acquired through daily clinical routine in the trauma surgery department. Several years are needed until experienced trauma surgeons are significantly differentiated from residents. Inter-reader agreement varies from 66\% among residents to 71\% among experienced trauma surgeons \cite{Zuckerman1996hip}. To reach a precise classification, medical students and young trauma surgeons rely on a second opinion to choose the adequate treatment option for the patient. Our work aims to provide support as a computer-aided diagnosis (CAD) system capable of classifying radiographs.

Convolutional neural networks (CNNs) are nowadays the model of predilection for CAD. They have been rapidly integrated in numerous medical applications \cite{Shin2016CNNCAD, Shen2017DLReview, Gibson2018NiftyNet, Bonavita2020Pulmonary, Bernard2018SegmCAD} due to their strong capacity to learn, directly from data, meaningful and hierarchical image representations. However, their feature extraction ability heavily depends not only on the optimization scheme but also on the training dataset. To be properly trained, CNNs need a large dataset representative of the population of interest \cite{Krizhevsky2012imagenet}.

In general, in medical image analysis tasks, acquiring reliable and clinically relevant annotated data remains a key challenge. Apart from the intra- or inter-expert disagreement, typically, manual annotations call for the time and effort of clinical experts. In addition, medical datasets usually suffer from class imbalance due to difficulties in collecting cases and the incidence of rare diseases. Finally, medical image data needs also dealing with proprietary and/or privacy concerns. As a result, these datasets generally exhibit three main challenging characteristics: (i) limited amounts of data, (ii) class-imbalance, and (iii) uncertain annotations. 

The most common approaches to alleviate these challenges have been transfer learning~\cite{Shin2016CNNCAD, zhou2017usingTL, Bonavita2020Pulmonary, HongShang2019TLSemi}, data augmentation~\cite{vasconcelos2017dataagu} and semi-supervised learning~\cite{HaiSu2019MeanTeachers, HongShang2019TLSemi}. More recently, the attention has been shifted towards bootstrapping or weighting strategies~\cite{roy2017error}, sample mining~\cite{Xue2019SampleMining}, active learning~\cite{Smailagic2018MedAL}, and curriculum learning~\cite{Tang2018AttentionGuidedCL,Yang2019SPBLSkin,Maicas2018likeRadiologists,Jimenez2019MedicalCurriculum}.

The underlying intuition of strategies such as reordering, sampling or weighting, is that they can significantly impact the optimization of CNNs during training. Towards this objective, we reunite and formulate the above curriculum learning (CL) strategies to improve the performance of fine-grained proximal femur fractures classification, by dealing with the lack of large annotated datasets, class imbalance, and annotation uncertainty. Inspired by the concept of curriculum in human learning, CL presents the training samples to the algorithm in a meaningful order (often by difficulty from ``easy'' to ``hard'') and has been shown to avoid bad local minima and lead to an improved generalization \cite{Bengio2009CL}.

Lately, training CNNs with ordered sequences has been shown to improve medical image segmentation by gradually increasing the context around the areas of interest \cite{Havaei2016, Jesson2018-CASED, Kervadec2019CurriculumSemi}. To the best of our knowledge, only few works have explored sample reordering for CAD with CNNS, for instance by extracting prior knowledge from radiology reports~\cite{Tang2018AttentionGuidedCL} or medical guidelines~\cite{Jimenez2019MedicalCurriculum}.

The ordering can be either fixed (\eg set heuristically by a ``teacher'' or domain-specific knowledge) or, in the absence of a-priori knowledge, a self-paced order~\cite{Kumar2010-selfpaced} derived from the algorithm's performance (\eg{}the loss). Our unified CL formulation encompasses both approaches. We address the lack of prior knowledge to design an ad-hoc curriculum, by providing a ranking criterion based on uncertainty modelling. By using uncertainty to define our ranking, the classifier favors samples that it has not yet properly learnt, thus guiding it to explore ``unseen'' parts of the input space. We present three manners to actually implement the curriculum data sequencing. The first one is based on reordering the training set. The second uses a sampling strategy, \ie selecting increasingly growing subsets. The last one employs a weighting scheme to give different importance to the training samples. 

To show the impact of our proposed method, we perform two types of experiments. First, on the challenging problem of multi-class classification of proximal femur fractures. This multi-class problem is inherently imbalanced, as the frequency of the classes reflects their incidence. Moreover, the adequate classification takes several years of daily clinical routine in the trauma surgery department, limiting the collection of annotations and leading to potentially noisy labels. Thus, to deepen the understanding of the method and to verify its effectiveness under these challenging data conditions, we design a series of experiments on the MNIST dataset, controlling the amount of data, class-imbalance, and label noise.

%%%%%%%%%%%%%%%%%%%%%%%%%%%%%%%%%%%%%%%%%%%%%%%%%%%%% Figure 1:
\begin{figure*}[t]
    \centering
    \includegraphics[width=0.8\textwidth]{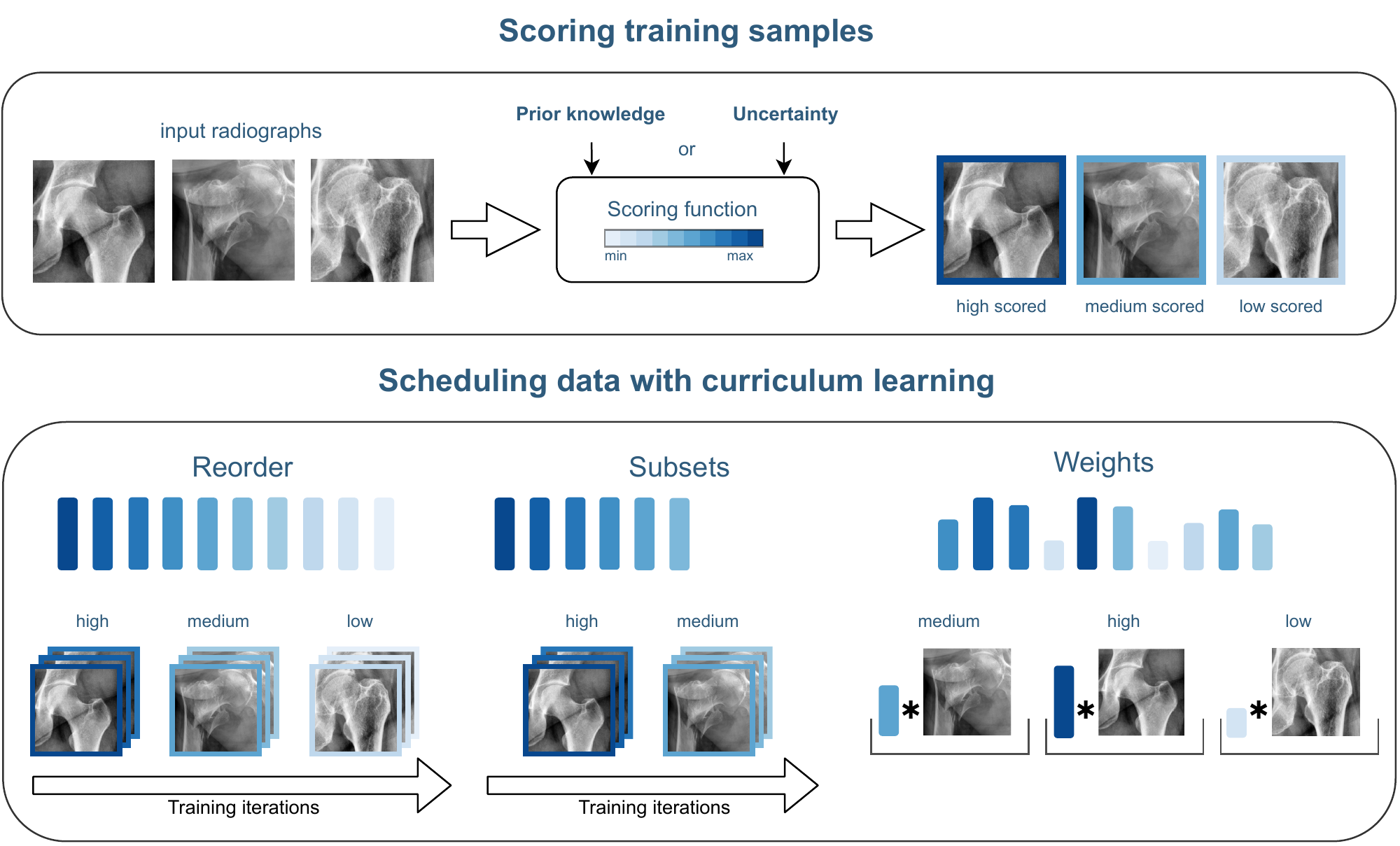}%
    \caption{ Training a CNN with CL. Top: training samples are scored according to prior knowledge or uncertainty. Bottom: three CL strategies are presented to automatically schedule the order and pace of the training samples for multi-class classification.} %
    \label{fig:abstract}%
\end{figure*}
%%%%%%%%%%%%%%%%%%%%%%%%%%%%%%%%%%%%%%%%%%%%%%%%%%%%%%%%%%%%%%%%%%%%%%

\paragraph*{\textbf{Contributions}} 
In this work, we propose three CL strategies to automatically schedule the order and pace of the training samples for an improved multi-class classification. Our contributions are:
\begin{itemize} 
    \item We identify common curriculum learning elements among different data scheduling strategies, and present them within a unified formulation.
    \item We propose two types of novel ranking functions guiding the prioritization of the training data.
    \item We leverage domain-specific clinical knowledge to define the first scoring function.
    \item In absence of domain knowledge, we propose to estimate the ranking of the training samples by dynamically quantifying the uncertainty of the model predictions.
    \item We validate our strategies on a clinical dataset for the multi-class classification of proximal femur fractures.
    \item With a controlled experimental setting, we confirm that our method is useful in reducing the classification error under limited amounts of data, imbalance in the class distribution, and unreliable annotations. We give recommendations about the best approaches for each scenario.
\end{itemize}

This paper is structured as follows. Section~\ref{sec:related} covers CL related works that are relevant for the design of data schedulers. In Section~\ref{sec:method}, the details of our proposed formulation are presented. Section~\ref{sec:experimental} describes the specifications of the experimental validation. Section~\ref{sec:results} shows the classification performance. Section~\ref{sec:discussion} discusses our findings, recommendations and future work. Finally, Section~\ref{sec:conclusions} summarizes our conclusions. 

\section{Related work} \label{sec:related}
Recently, CL, self-paced learning (SPL), active learning (AL) and selection strategies have been studied to improve CNN-based image classification performance. These methods rely on ranking the training samples according to some criterion. In the following, we highlight some works that employ the two criteria related to our method: (i) domain-specific prior knowledge and (ii) data and model uncertainty.

Prior knowledge is leveraged in \cite{Tang2018AttentionGuidedCL, Jimenez2019MedicalCurriculum, Yang2019SPBLSkin, Maicas2018likeRadiologists} to design a curriculum for classification. Yang~\etal~\cite{Yang2019SPBLSkin} exploited SPL to handle class-imbalance, by combining the number of samples in each class and the difficulty of the samples, which is derived from the loss. Tang~\etal~\cite{Tang2018AttentionGuidedCL} proposed to feed the images in order of difficulty based on severity-levels mined from radiology reports to improve the localization and classification of thoracic diseases. \JimSan~\etal~\cite{Jimenez2019MedicalCurriculum} exploited the knowledge of the inconsistencies in the annotations of multiple experts and medical decision trees, to design a medical-based deep curriculum that boosted the classification of proximal femur fractures. Trying to mimic the training of radiologists, Maicas~\etal~\cite{Maicas2018likeRadiologists} proposed to pretrain a CNN model with increasingly difficult tasks, before training for breast screening. The pretraining tasks were selected using teacher-student CL. In this work, we schedule our training data based on a scoring function that ranks the samples according to domain-specific prior knowledge or uncertainty. Different from previous works~\cite{Tang2018AttentionGuidedCL, Jimenez2019MedicalCurriculum, Maicas2018likeRadiologists}, which only considered reordering the training set, here we investigate two further curriculum strategies, namely, subset sampling and weighting. Furthermore, solely Yang~\etal~\cite{Yang2019SPBLSkin} targeted one of the mentioned data challenges: class-imbalance, whereas we investigate as well noisy labels and limited amounts of training data. 

The second criterion that we consider for defining a curriculum is uncertainty. The estimation of uncertainty provides a way of systematically defining the difficulty of the samples. Xue~\etal~\cite{Xue2019SampleMining} proposed online sample mining based on uncertainty to handle noisy labels in skin lesion classification. In their work, uncertainty is approximated through the classification loss. However, the most common methods for estimating classification uncertainty, in the context of deep learning, rely on Bayesian estimation theory, namely using Monte-Carlo (MC) dropout~\cite{Gal2016BayesianDropout}. Uncertainty is probably the most frequent criterion in AL selection strategies. Recently, Wu~\etal~\cite{Wu2018ALUncer} combined uncertainty together with image noise into their AL scheme to alleviate medical image annotation efforts. Uncertainty and label correlation are integrated in the sampling process to determine the most informative examples for annotation. AL pays attention to examples near the decision surface to infer their labels. Similarly, we aim to gradually move the classification decision border by adding examples of increasing ranking scores. We prioritize in our second scoring function the most representative samples, letting uncertainty guide their order, pace or weight. Although uncertainty has been used as sampling criterion for AL, we employ this information, for the first time, to rank and define our curriculum.

We validate our proposed curriculum strategies for the classification of proximal femur fractures. Whereas most of the previous work on femur fractures focuses on the binary fracture detection task~\cite{Badgeley2019, Cheng2019,Wang2019WeaklySupervised}, we target the more challenging  multi-class classification according to the AO standard~\cite{kazi2017AutomaticClassification, Jimenez2019MedicalCurriculum, JimenezSanchez2019IPCAI}.

Approaches to boost fracture classification accuracy comprise prior localization, transfer learning or medical knowledge. The localization of a region of interest before the classification of the full image has been studied either in a weakly-supervised  \cite{Wang2019WeaklySupervised, kazi2017AutomaticClassification} or in a supervised \cite{JimenezSanchez2019IPCAI} way. Knowledge transfer has been investigated across image domains, \ie using ImageNet dataset for pretraining \cite{Urakawa2019,Badgeley2019}, and across tasks, \ie training first on body part detection (easier task) and then focusing on the hip fracture detection \cite{Cheng2019}. Medical knowledge has been proposed to train a hierarchical cascade of classifiers \cite{tanzi2020hierarchical} or to schedule training data into a set of increasing difficulty \cite{Jimenez2019MedicalCurriculum}. Tanzi~\etal~\cite{tanzi2020hierarchical} relied on a cascade of classifiers. However, this kind of strategies are prone to propagate errors in multi-class classification. Furthermore, our CL approach does not rely on a complicated multistage scheme. We do not introduce any further complexity to the CNN.

In our previous work \cite{Jimenez2019MedicalCurriculum}, a series of heuristics, based on knowledge such as medical decision trees and inconsistencies in the annotations of multiple experts, were proposed as a scoring function to boost fracture classification performance. Here, we further propose two more strategies, and also provide an alternative mechanism to rank the samples, based on prediction uncertainty, in case prior knowledge is unavailable.

\section{Method} \label{sec:method}
Given a multi-class image classification task, where an image $x_i$ needs to be assigned to a discrete class label $y_i \in \{1, \dots, T\}$, our training set is defined as $X = \{(x_1,y_1), \ldots, (x_N,y_N)\}$. Assume a CNN model $h$ with parameters $\theta$ is trained with stochastic gradient descent (SGD). During training, samples are typically randomly ordered. Our goal is to instead schedule the order and pace of the training data presented to the optimizer to better exploit the available data and annotations, and thereby improve the classification performance.

To learn the best CNN model $h_{\theta^{*}}$ from the input data, a common choice is to use empirical risk minimization: 
\begin{align} \label{eq:erm}
\begin{split}
    \mathcal{L}(\theta) &= \tilde{\EX}[L_{\theta}] = \dfrac{1}{N}\sum_{i=1}^{N}L_{\theta}(x_i, y_i) \\
    \theta^* 
    &= \argmin_{\theta} \mathcal{L}(\theta)
\end{split}
\end{align}
where $\tilde{\EX}$ stands for the empirical expectation, $L_{\theta}$ is the loss function that measures the cost of predicting $h_{\theta}(x_i)$ when the correct label is $y_i$. 

Optimization is conducted with SGD for a total of $E$ epochs. Typically, the objective function $L_{\theta}$ is non-convex and is minimized in mini-batches of size $B$. Whereas convex learning is invariant to the order of sample presentation, CNNs are not. In the later case, the loss function usually presents a highly non-convex shape with many local minima, so the order of sample presentation affects learning, and thus, the final solution. It has been empirically shown that the variance in the direction of the gradient step defined by easier examples is significantly smaller than that defined by difficult ones, especially at the beginning of training \cite{Needell2016SGDRandomizsed, Weinshall2018CLbyTL}. This suggests that favoring the easier examples may increase the likelihood to escape the attraction basin of an initial poor local minimum. {Taking into account the mini-batches, we can rewrite} Eq.~\eqref{eq:erm} as:
\begin{equation} \label{eq:erm-batches}
    \mathcal{L}(\theta)=  \dfrac{1}{N}\sum_{j=1}^{N/B}\sum_{k=1}^{B} L_{\theta}(\hat{x}_{k,j}, \hat{y}_{k,j}),
\end{equation}
where $\hat{x}_{k,j}$ is the k-th sample in the j-th batch, $\hat{x}_{k,j} = x_{k + (j-1) \cdot B}$, and $\hat{y}_{k,j}$ is the corresponding label.

We propose to modify Eq.~\eqref{eq:erm-batches} to schedule the training data. To do so, first, we formalize two types of scoring functions to assign a priority level to each data sample. The scoring is defined in Subsection~\ref{subsec:scoring} either according to domain-specific prior knowledge or to the samples' uncertainty measured with MC dropout. Then, in Subsection~\ref{subsec:scheduler}, we introduce the different components required for reordering, pacing, and weighting the training data. Fig.~\ref{fig:abstract} provides an schematic illustration of the process for training a CNN with CL. Finally, we cover the implementation details of the three variants of our unified CL formulation in Subsection~\ref{subsec:framework}.

%% Figure:3 STRATEGIES FLOW diagram:
\begin{figure}[t]
    \centering
    \includegraphics[width=0.45\textwidth]{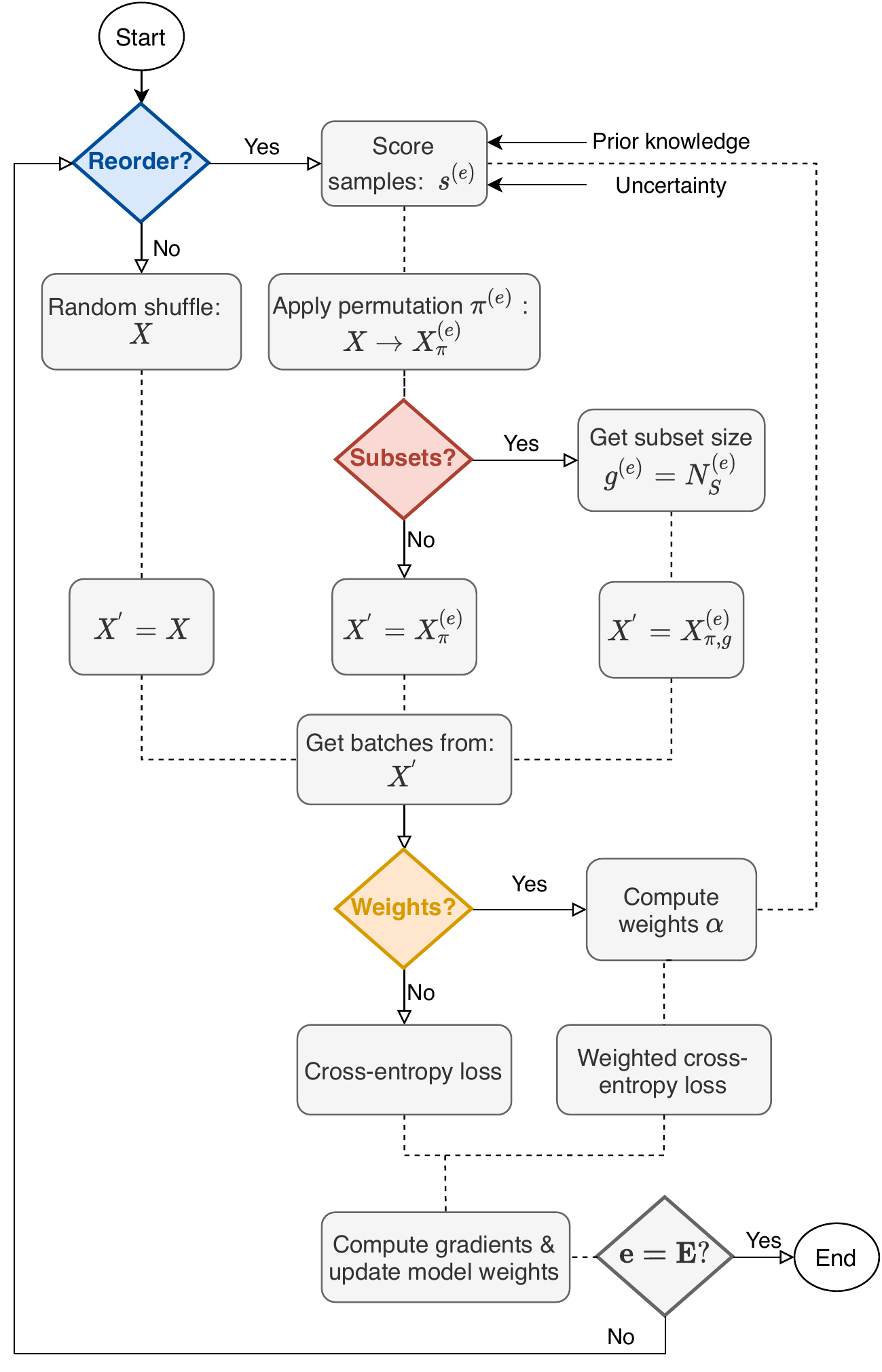}%
    \caption{ 
    Diagram illustrating the components of the proposed unified CL method reuniting the three scheduling strategies: reorder, subsets, and weights. Straight lines are employed after a Yes/No junction because the flow is split. Otherwise, dotted lines are employed when there is no split.}
    \label{fig:data-scheduler-flow}%
\end{figure}

\subsection{Scoring function definition} \label{subsec:scoring}
The key element of our approach is the definition of the scoring function $s$ or, equivalently, the curriculum probabilities $p$, which corresponds to normalized score function values. The formal definition of the curriculum probabilities is presented in Subsection~\ref{subsec:scheduler}. The curriculum allow us to sample the dataset and obtain a reordering function $\pi$ that schedules the training samples. In this subsection, we present two alternative scoring functions. The first one is static and based on some initial (domain) knowledge, as in classical CL~\cite{Bengio2009CL}. The second one is dynamic and based on the estimation of uncertainty, inspired by SPL~\cite{Kumar2010-selfpaced,Wang2019DynamicCL}.

\subsubsection{Prior knowledge} 
In this scenario, the initial scoring $s^{(0)}$ and, thus, curriculum probabilities $p^{(0)}$, are specified based on domain prior knowledge. We assume in this {variant} that the scoring values are defined per class:
\begin{equation} 
 s^{(0)}(\cdot, y=t) = \omega_t,    
\label{eq:PCL}
\end{equation}
where $t \in \{1, \dots, T\}$ serves as index of the classes. $\omega_t$ is defined specifically for each task (or dataset). Once that the scoring values have been initialized, they can be kept fixed or decayed towards a uniform distribution \cite{Bengio2009CL}. In either case, as the curriculum probabilities are predetermined a priori in Eq.~\eqref{eq:PCL}, we refer to this approach as static CL.

Prior knowledge can be obtained, for example, extracting keywords from medical reports~\cite{Tang2018AttentionGuidedCL}, based on the frequency of samples~\cite{Yang2019SPBLSkin, Jimenez2019MedicalCurriculum}, employing medical classification standards or quantifying inconsistencies in the annotations \cite{Jimenez2019MedicalCurriculum}. Specifically for this work, we define the initial probabilities for the proximal femur fracture images based on the Cohen's kappa coefficient~\cite{Hallgren2012kappa}. This statistic is used to measure the agreement of clinical experts on the classification between two readings. Basically, the kappa coefficient quantifies the ratio between the observed and chance agreement. To better understand and illustrate the potential of CL, we also analyze our method on MNIST dataset. In this case, we extract prior knowledge by ranking the per-class $F_1$-score performance after few epochs of training. The exact values used for our experiments are specified in Subsection~\ref{subsec:impl-details}. 

\subsubsection{Uncertainty estimation}
In absence of domain knowledge, we propose to estimate the priority of the training samples by dynamically quantifying the uncertainty of the model predictions. Uncertainty provides a way of systematically ranking the training samples based on the model's agreement on the predictions, with the benefit of not requiring any prior knowledge. At each epoch $e$, we compute the uncertainty in predicting a sample $x_k$, and use such uncertainty as its scoring value $s_k$. See Subsection~\ref{subsec:scheduler} for the definitions of $x_k$ and $s_k$. The goal is to emphasize samples with high information gain at early stages of training, \ie to rapidly reduce the error in highly-misleading samples.

To estimate the uncertainty of the model predictions, we employ MC dropout~\cite{Gal2016BayesianDropout}. In this training regime, each epoch includes two stages~\cite{Yingting2019CertaintyDriven}: uncertainty estimation and label prediction. In the uncertainty estimation stage, we perform $L$ stochastic forward passes on the model under random dropout. The $L$ estimators are used to measure the uncertainty of the output of the model. In the prediction stage, a single forward pass is performed. Then, the classification loss is used to measure the difference between the prediction and the label.

Let $\sigma \in \realnumbers^T$ be the (softmax) output of the CNN. This output represents the probability distribution of the predicted label over the set of the possible classes for sample $x$, \ie $P(y=t \mid x,\theta):=\sigma_t$. We measure uncertainty as the entropy~\cite{Shannon1948Mathematical} of the output distribution, \ie predictive entropy:
\begin{equation} \label{eq:entropy}
    H(y|x,\theta) = - \sum_{t=1}^{T} P(y=t \mid x, \theta) \cdot \log P(y=t \mid x, \theta).
\end{equation}
This measurement helps to discriminate points that are far from all training data, yet the model assigns high confident prediction (low predictive entropy). We aim to minimize the effect of these samples, with a small weight or bringing them at a later stage in training \cite{Smith2018understanding}.

The output distribution $P(y=t|x, \theta)$ can be approximated using MC integration:
\begin{equation} \label{eq:output-distribution} %  \dfrac{1}{K}
    \tilde{P}(y=t \mid x, \theta) = \dfrac{1}{L} \sum_{l=1}^{L} P(y=t \mid x, \theta_l),
\end{equation}
where $P(y=t \mid x, \theta_l)$ is the probability of input $x$ to take class $t$ with model parameters $\theta_l \sim q(\theta)$, with $q(\theta)$ being the (dropout) variational distribution. We set the scoring function to be the estimated predictive entropy, computed from the MC estimated output distribution $\tilde{\sigma}_t=\tilde{P}(y=t \mid x, \theta)$: 
\begin{equation}
    s = - \sum_{t=1}^{T} \tilde{\sigma}_{t} \cdot \log \tilde{\sigma}_{t}.
\end{equation}
By assigning low scoring values to predictions with low predictive entropy, we decrease the priority of samples with low information gain. Note that in contrast with Eq.~\eqref{eq:PCL}, here, the scoring elements $s_i$ are defined independently for each sample, and updated after each epoch. Only few works measure uncertainty while learning the classification task \cite{Ghesu2019ClassifUncert}. To the best of our knowledge, our proposed dynamic uncertainty-driven curriculum strategy is novel for CAD.

\subsection{Data scheduler} \label{subsec:scheduler}
In the following, we define the scheduling elements required for reordering and pacing our training data: a scoring function~$s$, curriculum probabilities~$p$, a permutation function~$\pi$, a pacing function~$g$, and a weighting function~$\alpha$. The data scheduler takes as input the training set $X$, the scoring and pacing functions, $s$ and $g$, respectively, and it outputs the reordered set/subset, partitioned in mini-batches. All components are updated at each epoch $e$.
\begin{itemize} 
    \item The \textit{scoring function} $s:\mathcal{X} \xrightarrow{} \realnumbers$ ranks the curriculum priority of each training pair. The curriculum priority can take various forms, such as difficulty or prediction disagreement. %Then, 
    An example $(x_i, y_i)$ has higher priority than example $(x_j, y_j)$ if $s(x_i, y_i) > s(x_j, y_j)$. We define $s_i = s(x_i, y_i)$ and, in an abuse of notation, use $s$ to denote both the scoring function and the vector $(s_1, \ldots, s_N)$. 
    \item The \textit{curriculum probabilities} $p$ are obtained by normalizing the score function values (while preserving the order and ensuring they add up to 1). 
    For example, one can choose $p_i = s_i/||s||_1$, assuming $s_i \geq 0$. 
    A pair $(x_i, y_i)$ is more likely to be presented earlier to the optimizer than a pair $(x_j, y_j)$ if $p_i > p_j$. 
    \item The \textit{reordering function} $\mathbf{\pi} : [1,\ldots, N] \xrightarrow{} [1,\ldots, N] $ is a permutation. It is determined by resampling without replacement $X$ according to the curriculum probabilities $p$. 
    \item The \textit{pacing function} $g : \mathbb{N} \xrightarrow{} \mathbb{N}$ controls the learning speed by presenting growing subsets of data. The batch size $B$ is kept fixed. The non-decreasing mapping $g$ determines the subset size $N_S \leq N$ at each training epoch $e$, \ie $g(e)=N_S^{(e)}$. 
    \item The \textit{weighting function} $\alpha:\mathcal{X} \xrightarrow{} \realnumbers$ favors the samples that have higher priority according to the curriculum probabilities. These per-sample weights are applied directly to the classification loss. 
\end{itemize}

Taking into account the scheduling elements introduced, we can rewrite the optimization loss at epoch $e$ as:
\begin{equation}\label{eq:framework}
    \mathcal{L}^{(e)}_{\theta} =  \dfrac{1}{N_S^{(e)}} \sum_{j=1}^{N_S^{(e)}/B}\sum_{k=1}^{B} \hat{\alpha}_{k,j}^{(e)} \cdot L_{\theta}(\hat{x}_{k,j}^{(e)}, \hat{y}_{k,j}^{(e)}),
\end{equation}
where $\hat{x}_{k,j}^{(e)} = x_{\pi^{(e)}{(k + (j-1) \cdot B)}}$ corresponds to the k-th sample from the j-th batch at epoch \textit{e} after reordering $\pi$. The same relation follows for its corresponding label and weight, $\hat{y}_{k,j}^{(e)}$ and $\hat{\alpha}_{k,j}^{(e)}$, respectively. We will drop superscript $(e)$ when no confusion arises. Also, we simplify notation and use $x_k$ (and $y_k$, $\alpha_k$) to refer to a given (already reordered) sample (and label, weight). This equation encompasses the three main curriculum strategies from the literature: reordering, increasing subsets, and weighting. 

\subsection{Scheduling data with curriculum learning} \label{subsec:framework}
In practice, any curriculum is implemented by assigning a predefined or estimated probability $p_i$ to each training pair $(x_i,y_i)$, as described in Subsection~\ref{subsec:scoring}. Fig.~\ref{fig:data-scheduler-flow} visualizes the data flow in the different scheduling strategies, each of them being depicted by a diamond shape: reorder, subsets, and weights. The scoring function $s$ and curriculum probabilities $p$ are common to the three scheduling approaches, whereas the reordering function $\pi$ is used in the reorder and subset strategies.

The first mechanism, \textit{reorder}, presents the samples to the optimizer in a ``smart'' probabilistic order, instead of the typical random permutation. This strategy aims to deal with low-priority cases at a later stage of training~\cite{Bengio2009CL, Hacohen2019PowerCL, Jimenez2019MedicalCurriculum}. At the beginning of every epoch e, the training set $X = \{(x_1,y_1), \ldots, (x_N,y_N)\}$ is permuted to $X_{\pi}^{(e)} = \{(x_{\pi^{(e)}(1)},y_{\pi^{(e)}(1)}), \ldots, (x_{\pi^{(e)}(N)},y_{\pi^{(e)}(N)})\}$ using the reordering function $\pi^{(e)}$. This mapping results from sampling the training set according to the curriculum probabilities $p^{(e)}$ at the current epoch $e$. Mini-batches are formed from $X_{\pi}^{(e)}$.

The second method, \textit{subsets}, builds upon the reordered training set and selects gradually increasing subsets at every epoch. The purpose is to reduce the effect of outliers at the beginning of training~\cite{Hacohen2019PowerCL, Xue2019SampleMining, Wang2019DynamicCL}. Mini-batches are obtained from $X^{(e)}_{\pi, g} \subseteq X$, where $X^{(e)}_{\pi, g}$ are the first $N_S^{(e)}$ pairs of $X^{(e)}_{\pi}$. The subset size at every epoch $N_S^{(e)}$ is determined by the pacing function $g$. For simplicity, in our experiments we choose $g$ to be a staircase function:
 \begin{equation} \label{eq:grow} 
 g(e) = N_S^{(e)} = \left\{ \begin{array}{lcc}
 N_S^{(0)} + e \cdot \Delta &  if & 1 \leq e < E_S \\ 
 \\ N  & if & e \geq E_S
    \end{array} \right.
\end{equation}
where $\Delta = (N-N_S^{(0)})/E_S $, $N_S^{(0)}$ is a predefined initial subset size, and $E_S$ is the number of epochs before considering the whole training set.

A counter $\tau_i$ is introduced to track the selected pairs. Their scoring vector is decreased, thus favoring new pairs in the subsequent epoch. We choose to update the scoring vector using an exponential decay: 
\begin{equation}
s_i^{(e)} = s_i^{(e-1)} \cdot \exp({-\tau_i^{2}}/{10}) \quad e=1, \dots, E.
\label{eq:decay} \\
\end{equation}

The third approach, \textit{weights}, assigns scalar weights to training samples based on their curriculum probabilities~\cite{Wang2019DynamicCL}. We propose to weight the classification loss $L_{\theta}$ of each training sample in Eq.~\eqref{eq:framework}, in the form of a weighted cross-entropy loss. The role of the weights is to decrease the contribution to the classification loss of samples with low priority. We choose the weights $\hat{\alpha}_{k,j}$ to correspond to a per-batch normalization of the curriculum probabilities:
\begin{equation}
    \hat{\alpha}_{k,j}^{(e)} = \dfrac{ p_{k+(j-1) \cdot B}^{(e)}}{\max \limits_{m} p_{m+(j-1) \cdot B}^{(e)}} = 
    \dfrac{\hat{p}_{k,j}^{(e)}}{\max \limits_{m} \hat{p}_{m,j}^{(e)}}.
\end{equation} 

When the curriculum is driven by uncertainty, the resulting approach is similar to boosting~\cite{Freund1999Boosting}. In the boosting method, misclassified examples are given a higher weight than correctly classified ones. This is known as ``re-weighting''. Following the same principle, we use the uncertainty at every epoch, in our curriculum data scheduler, to update the values of the weights.

%%%%%%%%%%%%%%%%%%%%%%%%%%%%%%%%%%%%%%%%%%%%%%%%%%%%% Figure
% MNIST Curriculum learning with prior knowledge.
\begin{figure*}[t]
    \centering
    \subfloat[Prior knowledge-driven CL]{{\includegraphics[width=0.43\textwidth]{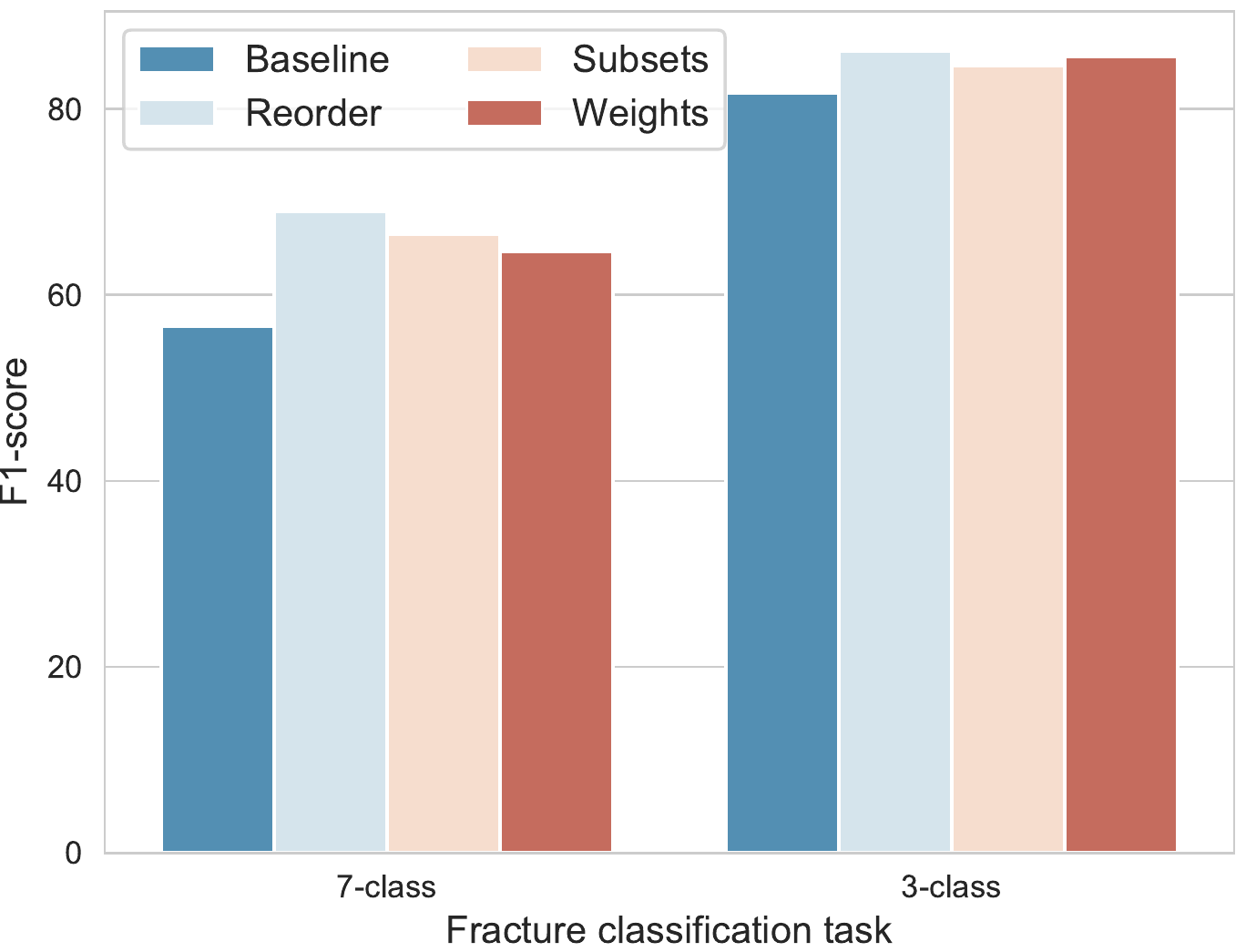} }}%
    \qquad
    \subfloat[Uncertainty-driven CL.]{{\includegraphics[width=0.43\textwidth]{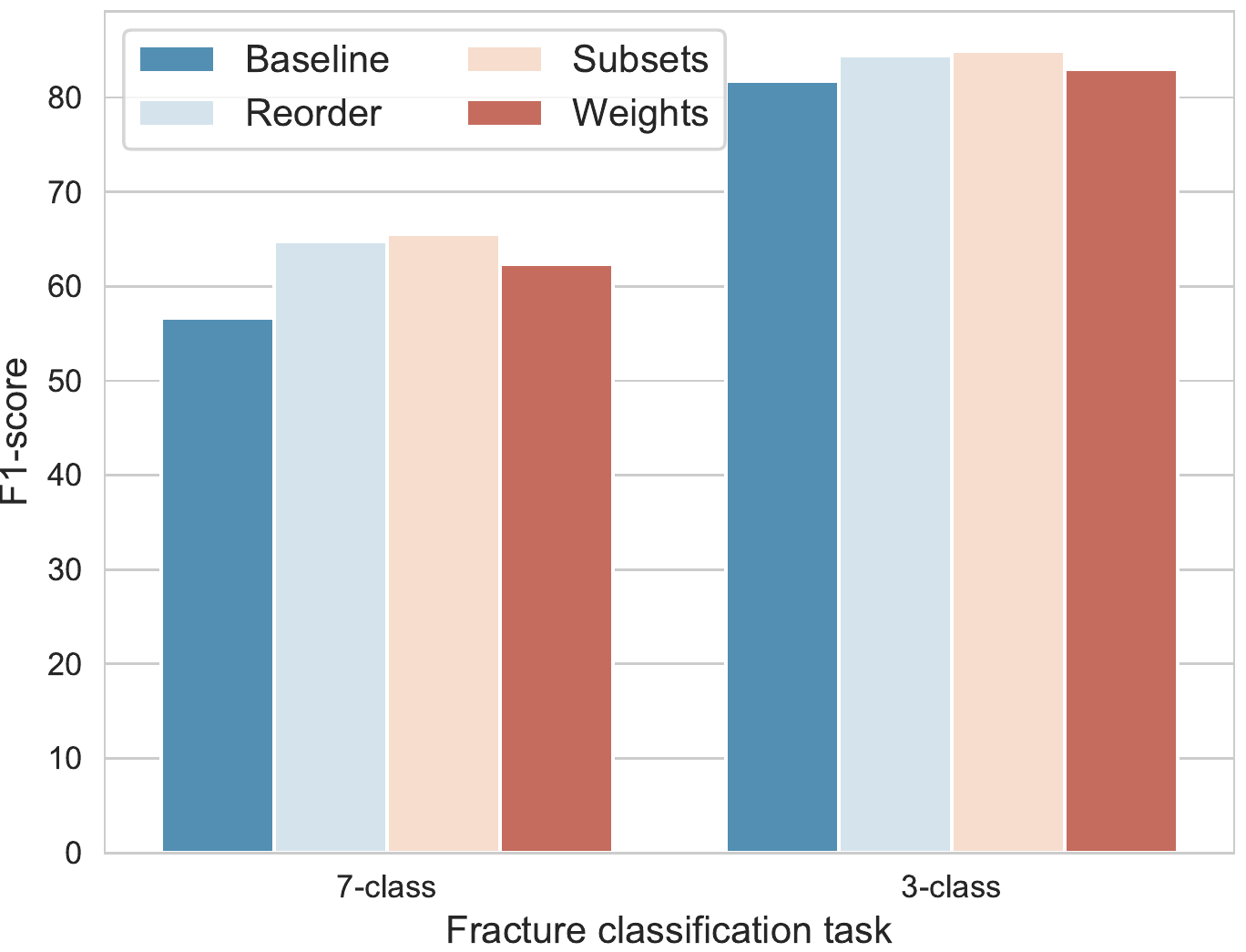} }}%
    \caption{F1-score for multi-class fracture classification. The proposed curriculum method improves the classification performance in all variants.}%
    \label{fig:barplotFract}%
\end{figure*}
%%%%%%%%%%%%%%%%%%%%%%%%%%%%%%%%%%%%%%%%%%%%%%%%%%%%%%%%%%%%%%%%%%%%%%

%%%%%%%%%%%%%%%%%%%%%%%%%%%%%%%%%%%%%%%%%%%%%%%%%%%%% Figure
% MNIST Curriculum learning with prior knowledge.
\begin{figure*}[t]
    \centering
    \subfloat[Prior knowledge-driven CL]{{\includegraphics[width=0.43\textwidth]{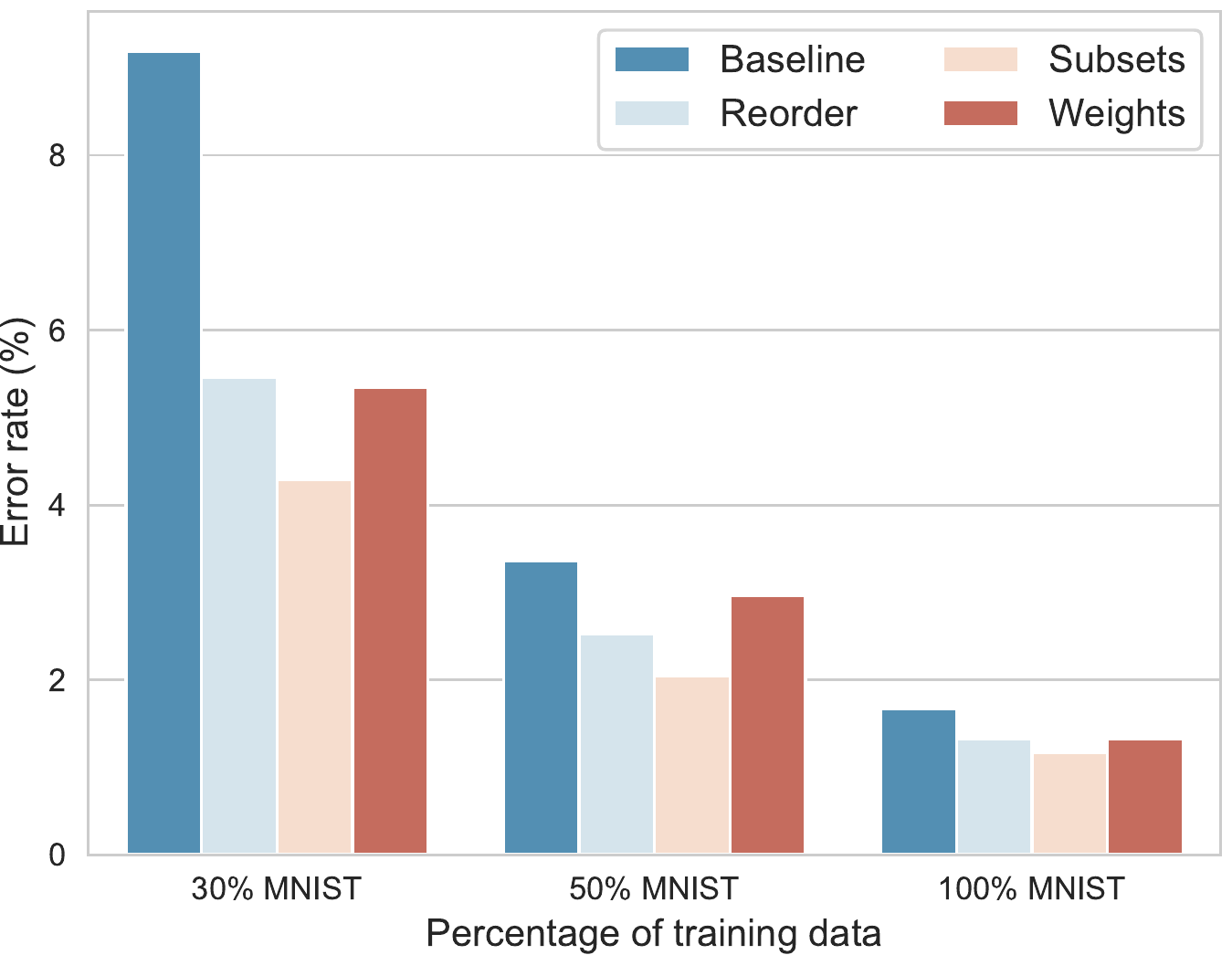} }}%
    \qquad
    \subfloat[Uncertainty-driven CL.]{{\includegraphics[width=0.43\textwidth]{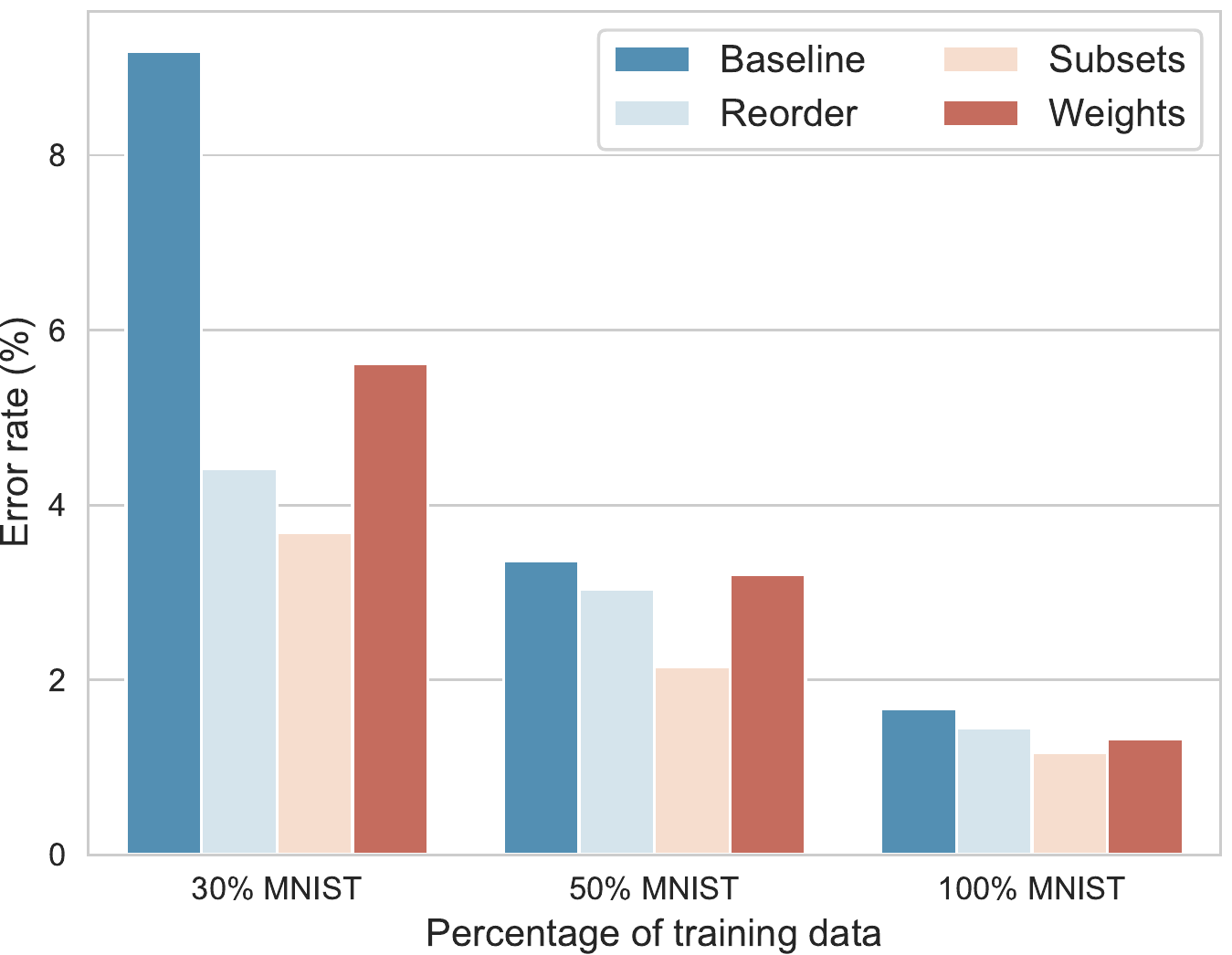} }}%
    \caption{Classification performance for digit recognition. The proposed curriculum method reduces the baseline error rate in all variants under limited amounts of data.}%
    \label{fig:CLandSPL}%
\end{figure*}
%%%%%%%%%%%%%%%%%%%%%%%%%%%%%%%%%%%%%%%%%%%%%%%%%%%%%%%%%%%%%%%%%%%%%%

\section{Experimental validation} \label{sec:experimental}
In order to validate the positive effect of data scheduling on the classification performance, we perform experiments on two types of image databases: (i) a real in-house dataset of a moderate size and naturally suffering from imbalance and noisy labels, and (ii) the MNIST dataset. The second one is used for additional analysis under controlled experiments to further illustrate the potential of CL.

\subsection{Datasets}
\paragraph*{\textbf{Proximal femur fractures}} 
Our clinical dataset consists of anonymized X-rays of the hip and pelvis collected at the trauma surgery department of the Rechts der Isar Hospital in Munich. Images of $2500 \times 2048$ pixels were gathered from a group of 780 patients. Each patient study contained one or two radiographs. Most of the images were Anterior-Posterior (A-P), only 4\% were side view. The collection of these radiographs was approved by the ethical committee of the Faculty of Medicine from the Technical University of Munich, under the number 409/15 S. The dataset consists of 327 type-A, 453 type-B fractures and 567 non-fracture cases. Class labels were assigned by clinical experts according to the AO classification standard \cite{Kellam2018}. Each type of fracture is further divided into 3 subclasses depending on the morphology and number of fragments of the fracture, see Fig.~\ref{fig:ao}. Subtypes of the fracture classes are highly unbalanced, reflecting the incidence of the different fracture types. In particular, the number of images for the subclasses is as follows: type-A (114, 197, 16), and type-B (79, 241, 133). Clinicians also provided square bounding box annotations containing the head and neck of the femur. We leveraged these annotations, cropped and resized the image to $224 \times 224$ pixels. The dataset was split patient-wise into three parts with the ratio 70\%:10\%:20\% to build respectively the training, validation and test sets. We evaluate the classification performance of the 3-class (type-A or type-B and non-fracture) and 7-class (fracture subtypes and non-fracture) classification tasks. The train, validation and test distributions were balanced between fracture type-A, type-B, and non-fracture cases. To achieve an equal proportion of subtype representation (of approximately 12\%), data augmentation techniques were used. Specifically, techniques such as translation, scaling and rotation were combined.

\paragraph*{\textbf{MNIST}} 
The MNIST handwritten digit database is publicly available\footnote{http://yann.lecun.com/exdb/mnist/}. It has a training set of 50000 examples and a validation and test sets of 10000 examples each. Classes are equally represented. 

\subsection{Experimental Setting}
We perform a comparative evaluation of the classification task with five series of experiments. Our method is contrasted against its ``anti-curriculum'' approach, \ie the curriculum probabilities are complemented so that training samples follow the reverse order, ``random'' criterion, \ie the curriculum probabilities are assigned randomly, and the ``baseline'' model. The baseline model does not consider any data scheduling elements, and it is trained on randomly shuffled versions of the whole training set.

In the first series of experiments, we examine the performance of our method driven by prior knowledge. In the second series, we consider the use of uncertainty to overcome the lack of prior knowledge. Our clinical dataset inherently suffers from class-imbalance, unreliable annotations and a limited size. The 7-class discrimination task is challenging as reflected by i) the existing intra- and inter-expert agreement (66\% among residents \vs 71\% among experienced trauma surgeons); and ii) the long and shallow learning curve of young trauma surgery residents who acquire the classification skills during the daily routine. For the remaining experiments, we employ MNIST, as a controlled environment, to investigate such challenging scenarios. In the third series, we evaluate the classification performance when training with limited amounts of data. In the fourth series, we present the results that deal with class-imbalance. Finally, in our last series of experiments, we discuss and show the performance under the presence of label noise.

\subsection{\textbf{Implementation details}} \label{subsec:impl-details}
\begin{figure}[ht]
    \centering
    \includegraphics[width=0.5\textwidth]{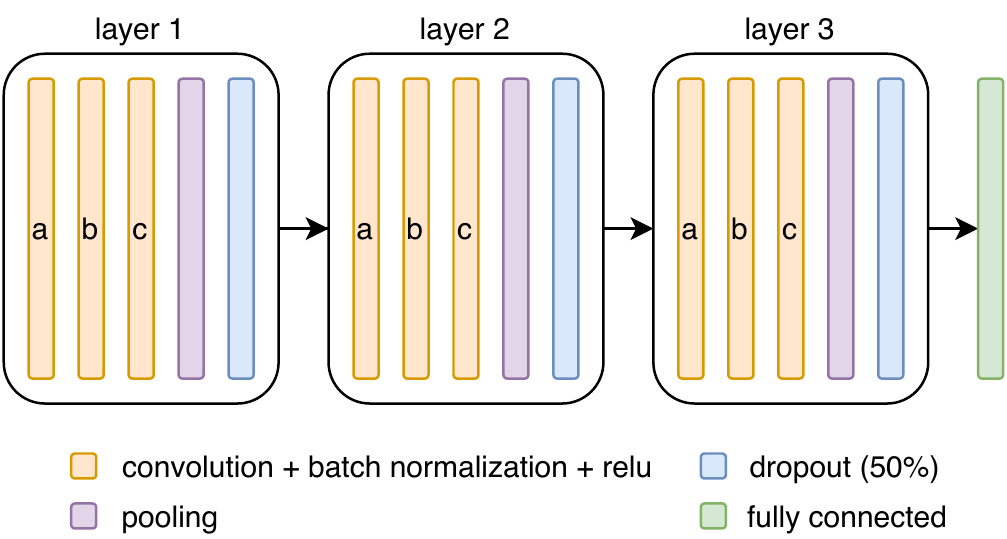}%
    \caption{Network architecture employed for the experiments with the MNIST dataset.}
    \label{fig:mnist-network}%
\end{figure}

\paragraph*{\textbf{Architectures and optimization hyperparameters}}
We train our models 10 times for 30 epochs, with an early stopping criterion of no improvement in the validation set for 20 epochs. For the digit recognition task, we use an upgraded ConvPool-CNN-C~\cite{Springenberg2014striving} proposed by \cite{Laine2017}, illustrated in Fig.~\ref{fig:mnist-network} of Suppl. Material. This architecture replaces pooling layers by convolutional layers with a stride of two. Besides, the small convolutional kernels greatly reduce the number of parameters of the network. It yielded competitive performance on several object recognition datasets (CIFAR-10, CIFAR-100, ImageNet). For the fracture classification, we deploy a ResNet-50 \cite{He2015-ResNet} pretrained on the ImageNet dataset, on account of the limited size of our dataset and the benefits of transfer learning \cite{Shin2016CNNCAD, HongShang2019TLSemi}. We limit our evaluation to those two CNNs, since Weinshall~\etal~\cite{Weinshall2018CLbyTL} reported that CL lead to an improved generalization performance with both `small' and `large' architectures. For both architectures, we use a mini-batch size of 64, an initial learning rate of $1\mathrm{e}{-3}$, and a dropout rate for the fully connected layer of 0.9 (0.7 for uncertainty estimation). Our ResNet-50 is trained with SGD and a momentum of 0.9. The learning rate is decayed by a factor of 10 every 10 epochs. ConvPool-CNN-C is trained with Adam. For the weighting strategy, since the batch size is directly related to the computation of the sample weights, we evaluated different batch sizes (16, 32, and 64). We found that the curriculum is robust, achieving the lowest standard deviation for $B=64$ (see Table~\ref{table:weights-bsize} in Suppl. Material). For the subsets strategy, we choose as hyperparameters: the warm-up epochs $E_S=10$ and the initial subset size $N_S^{(0)}$ to 25\% of the training data size at each scenario. We evaluated several warm-up epochs $E_S=\{5, 10, 20\}$ and sizes for the initial subset $N_S^{(0)}= \{25\%, 40\%\}$. Results for the different configurations were comparable (see Tables~\ref{table:subsets-init}-\ref{table:subsets-epochs} in Suppl. Material). 

\paragraph*{\textbf{Prior knowledge}}
\begin{itemize}
\item Proximal femur fractures. In this setting, we leverage, as prior knowledge the intra-reader agreement from a committee of experts: a trauma surgery attendant with one year experience, a trauma surgery attending and a senior radiologist. The scoring values for the seven classes are the following:
    \begin{equation}
        \omega = (0.69, 0.56, 0.62, 0.60, 0.56, 0.38, 0.92).
    \end{equation}
These values correspond to the multi-read kappa agreement described in Results section  \cite{JimenezSanchezTowardsCAD}.
\item MNIST. In absence of domain-specific knowledge, a CNN is trained for 5 epochs. After observing the $F_1$-score of each of the classes, weights are assigned, by ranking the classes from easiest (highest $F_1$-score) to hardest (lowest $F_1$-score). Then, training is restarted from scratch using these particular weights. We specify the values for the experiments with limited amounts of data $\omega_{limited}$, under class-imbalance $\omega_{imbalance}$, and with noisy labels  $\omega_{noise}$:
    \begin{align}
        \omega_{limited} &= (7, 10, 5, 4, 9, 1, 8, 6, 2, 3) \\
        \omega_{imbalance} &= (3, 10, 7, 8, 5, 6, 9, 4, 1, 2) \\
        \omega_{noise} &= (8, 10, 9, 7, 5, 1, 2, 3, 4, 6).
    \end{align}
\end{itemize}

\newpage
\section{Results} \label{sec:results}
\subsection{\textbf{Prior knowledge-driven CL}}
We evaluated the performance of the classifier with our data scheduler and verified that establishing a curriculum based on prior knowledge is a good and suitable option to improve classification performance. 
Results for proximal femur fracture are summarized in Table~\ref{table:fractures}-top, and for digit recognition in Table~\ref{table:mnist}-top. We found that the three variants helped to improve the performance of the two datasets. In contrast with the anti-CL approach, accuracy was increased with respect to the baseline.

For MNIST, we found that training starting with an easy subset, and gradually increasing the subset by adding more difficult samples was the best strategy for the three scenarios as shown in Fig.~\ref{fig:CLandSPL}-a. A comparable improvement with respect to the baseline was found when the decay of Eq.~\eqref{eq:decay}  was introduced in reorder strategy and sampling with replacement was performed instead.

For fracture classification, the $F_1$-score for 7-class was improved up to 15\% compared to the baseline (see Table~\ref{table:fractures} and Fig.~\ref{fig:barplotFract}). This score is comparable to {state-of-the-art results~\cite{Jimenez2019MedicalCurriculum}} and experienced trauma surgeons~\cite{vanEmbden2010}. Although related works report results for binary classification (fracture/no fracture), we are not aware of other teams doing fine-grained multi-class fracture classification. In this case, the best method was reordering the whole training set. We hypothesize that by reordering, we improve diversity by including the more challenging fine-grained fractures classification task than employing subsets of the data. Furthermore, as specified in Subsection~\ref{subsec:impl-details}, the CNN for fracture classification was pretrained, whereas for digit recognition the CNN was trained from scratch. From the results in Table~\ref{table:fractures}, we can say that our method is compatible with transfer learning. 
 
\subsection{\textbf{Uncertainty-driven CL}} 
Here, assuming lack of prior knowledge, we confirmed that uncertainty estimation can guide the data scheduling. Results are presented in Table~\ref{table:fractures}-bottom and Table~\ref{table:mnist}-bottom for fractures and MNIST, respectively. For the fine-grained 7-class proximal femur fractures classification, the $F_1$-score was improved up to 16\% compared to the baseline. In this case, we found that weighting the samples was not as beneficial as reordering or sampling subsets.  For digit classification, the error rate was reduced up to 30\%, see Fig.~\ref{fig:CLandSPL}-b. Anti-CL leading to a better performance than the baseline is a behaviour also reported in \cite{Bengio2009CL}. Furthermore, we found that this behaviour was sporadic and not statistically significant with respect to the baseline, whereas the CL approach was consistent and statistically significant.

Table~\ref{table:fractures-classes} presents an analysis of the classification performance per class. We found a uniform improvement within each class when using our CL strategies with respect to the Baseline. Furthermore, not all classes are equally improved. For example, the easiest `Normal' class is only slightly improved by a maximum of 6\%, whereas more difficult classes like A1 or B2 are improved by 30\% and 39\%, respectively. The difference for A3 might not be significant due to the limited number of samples in the test set for this class.

\subsection{\textbf{Limited amounts of data}} 
Table~\ref{table:mnist} shows the digit recognition performance when restricting the amount of training data to 30\% and 50\%. When employing our curriculum framework, the error rate for digit classification is reduced in all cases. We found that employing subsets in the first epochs based on uncertainty was the best strategy. Moreover, the effect of our curriculum approach was more evident on the more challenging scenario. The error rate was reduced by up to 59\% training with only only 30\% of the data. Interestingly, we found that when training with only 30\% of data, the use of random subsets also reduced the error rate. This behaviour goes along with some findings about training with partial data \cite{Nur2018RandomGrowing}. 

\subsection{\textbf{Class-imbalance}} We evaluated our proposed curriculum method in a controlled experiment under class-imbalance with the MNIST dataset. Specifically, the number of examples of two classes (digits $1$ and $7$) are limited to 30\% of the available cases. Results in Table~\ref{table:imbalance-noise} show that our approach can cope with class-imbalance and improved over the baseline result. Similar to the experiment with limited amount of data, the use of high-priority subsets, selected based on prior knowledge or uncertainty, was the best approach. The subsets approach reduced the error rate from 2.53\% to 1.79\%.

\subsection{\textbf{Noisy labels}} 
Using MNIST and a controlled setting, we corrupted a randomly selected 30\% of random training labels by assigning to them the subsequent label digit, \ie zeroes become ones, ones become twos, \etc Table~\ref{table:imbalance-noise} reports the mean error rate (\%) when evaluating the digit classification. We found that all the variants of our unified CL framework were effective to deal with noisy labels and beat the baseline. In this case, prior knowledge was not as beneficial as the estimation of model prediction uncertainty. The best variant was using uncertainty to weight the classification loss, reducing the error rate by 43\%. The fact that uncertainty performed better than prior knowledge was expected, since noise may affect individual samples and not entire classes. It is more reasonable to use a scoring function that independently affects the samples. Moreover, although reordering and subsets presented all the samples at convergence, weighting seemed to be the only strategy to remove or reduce the influence of the flawed labels.

%%%%%%%%%%%%%%%%%%%%%%%%%%%%%%%%%%%%%%%%%%%%%%%%%%%%%%%%%%%%%%%%%%%%%%
%% Table: Fractures
\begin{table*}[ht]  % bht
\centering
\ra{1.5} 
\caption{Fracture classification results over 10 runs: mean $F_1$-score. The highlighted indices in bold correspond to the best metric per curriculum method. The underlined values correspond to the best metric per scenario, \ie 3-class (type-A or type-B and non-fracture) and 7-class (fracture subtypes and non-fracture) classification. Statistical significance with respect to baseline is marked with *.}
\begin{tabular}{@{}l|c|cc|ccc|ccc@{}} \toprule
\multirow{2}{*}{\textbf{Prior knowledge}} &
      \phantom{Baseline} &
      \multicolumn{2}{c|}{\textbf{Reorder}} &
      \multicolumn{3}{c|}{\textbf{Subsets}} &
      \multicolumn{3}{c}{\textbf{Weights}} \\
 & Baseline & Anti-CL & CL & Random & Anti-CL & CL & Random & Anti-CL & CL \\
\midrule
7-class & 56.62 & 34.56 & \underline{\textbf{68.93*}} & 58.90 & 50.89 & \textbf{66.50*} & 58.26 & 55.20 & \textbf{64.65*} \\ 
3-class & 81.71 & 60.46 & \underline{\textbf{86.23*}} & 80.82 & 75.64 & \textbf{84.69*} & 80.66 & 75.33 & \textbf{85.66*} \\
\midrule \midrule
\multirow{2}{*}{\textbf{Uncertainty}} &
      \phantom{Baseline} &
      \multicolumn{2}{c|}{\textbf{Reorder}} &
      \multicolumn{3}{c|}{\textbf{Subsets}} &
      \multicolumn{3}{c}{\textbf{Weights}} \\
 & Baseline & Anti-CL & CL & Random & Anti-CL & CL & Random & Anti-CL & CL \\
\midrule
7-class & 56.62 & 61.29 & \textbf{64.70*} & 58.90 & 62.06 & \underline{\textbf{65.51*}} & 58.26 & 58.29 & \textbf{62.29*} \\ 
3-class & 81.71 & 82.48 & \textbf{84.38*} & 80.82 & 82.79 & \underline{\textbf{84.90*}} & 80.66 & 82.69 & \textbf{82.96*} \\
\bottomrule
\end{tabular}
\label{table:fractures}
\vspace{-1em}
\end{table*}
%%%%%%%%%%%%%%%%%%%%%%%%%%%%%%%%%%%%%%%%%%%%%%%%%%%%%%%%%%%%%%%%%%%%%%

%% Table
%%%%%%%%%%%%%%%%%%%%%%%%%%%%%%%%%%%%%%%%%%%%%%%%%%%%%%%%%%%%%%%%%%%%%%
%% Table: MNIST (static & dynamic)
\begin{table*}[t]  % bht
\centering
\ra{1.5} 
\caption{Digit classification results over 10 runs: mean error rate (\%). The highlighted values in bold correspond to the best metric per curriculum method. The underlined values correspond to the best metric per scenario, \ie percentage of data. Statistical significance with respect to baseline is marked with *.}
\begin{tabular}{@{}l|c|cc|ccc|ccc@{}} \toprule
\multirow{2}{*}{\textbf{Prior knowledge}} &
      \phantom{Baseline} &
      \multicolumn{2}{c|}{\textbf{Reorder}} &
      \multicolumn{3}{c|}{\textbf{Subsets}} &
      \multicolumn{3}{c}{\textbf{Weights}} \\
 & Baseline & Anti-CL & CL & Random & Anti-CL & CL & Random & Anti-CL & CL \\
\midrule
30\% MNIST & 9.19 & 9.28 & \textbf{5.46*} & 5.60 & 13.17 & \underline{\textbf{4.29*}} & 8.01 & 5.78 & \textbf{5.35*} \\ 
50\% MNIST & 3.36 & 5.21 & \textbf{2.53*} & 3.96 & 4.21 & \underline{\textbf{2.05*}} & 4.10 & 4.11 & \textbf{2.96*} \\ 
100\% MNIST & 1.67 & 2.53 & \textbf{1.32*} & 1.96 & 1.78 &
\underline{\textbf{1.17*}} & 1.98 & 1.79 & \textbf{1.32*} \\
\midrule \midrule
\multirow{2}{*}{\textbf{Uncertainty}} &
      \phantom{Baseline} &
      \multicolumn{2}{c|}{\textbf{Reorder}} &
      \multicolumn{3}{c|}{\textbf{Subsets}} &
      \multicolumn{3}{c}{\textbf{Weights}} \\
 & Baseline & Anti-CL & CL & Random & Anti-CL & CL & Random & Anti-CL & CL \\
\midrule
30\% MNIST & 9.19 & 8.94 & \textbf{4.42*} & 5.60 & 8.85 & 
\underline{\textbf{3.69*}} & 8.01 & 8.50 & \textbf{5.62*} \\ 
50\% MNIST & 3.36 & 3.23 & \textbf{3.04} & 3.96 & 4.21 & \underline{\textbf{2.15*}} & 4.10 & 4.77 & \textbf{3.21} \\ 
100\% MNIST & 1.67 & 2.29 & \textbf{1.45} & 1.81 & 2.02 & \underline{\textbf{1.17*}} & 1.99 & 1.66 & \textbf{1.33*} \\
\bottomrule
\end{tabular}
\label{table:mnist}
\vspace{-1em}
\end{table*}
%%%%%%%%%%%%%%%%%%%%%%%%%%%%%%%%%%%%%%%%%%%%%%%%%%%%%%%%%%%%%%%%%%%%%%

%%%%%%%%%%%%%%%%%%%%%%%%%%%%%%%%%%%%%%%%%%%%%%%%%%%%%%%%%%%%%%%%%%%%%%
%% Table: Fractures, improvement per class
\begin{table*}[t]  % bht
\centering
\ra{1.5}  
\caption{ Relative per-class $F_1$-score with respect to the baseline. Up arrows indicate improvement, down arrows stand for degradation of performance. \\The highlighted values in bold correspond to the best strategy per class. }
\begin{tabular}{@{}l|ccccccc|c@{}} \toprule
Method & A1 & A2 & A3 & B1 & B2 & B3 & Normal & Avg. \\ 
\midrule
Baseline & 29.50 & 61.40 & 25.20 & 52.80 & 29.20 & 48.70 & 84.80 & 57.72 \\
\midrule
Reorder (Prior K.)& \textbf{\textuparrow 30\%} & \textbf{\textuparrow 21\%} & \textuparrow 17\% & \textuparrow 22\% & \textbf{\textuparrow 39\%} & \textbf{\textuparrow 18\%} & \textbf{\textuparrow 6\%} & \textbf{\textuparrow 16\%} \\
Subsets (Prior K.) & \textuparrow 11\% & \textuparrow 15\% & \textuparrow 0\% & \textbf{\textuparrow 23\%} & \textuparrow 30\% & \textbf{\textuparrow 18\%} & \textbf{\textuparrow 6\%} & \textuparrow 13\% \\
Weights (Prior K.) & \textuparrow 22\% & \textuparrow 15\% & \textuparrow 23\% & \textuparrow 11\% & \textuparrow 17\% & \textuparrow 4\% & \textuparrow 5\% & \textuparrow 11\% \\
\midrule
Reorder (Uncertainty) & \textuparrow 8\% & \textuparrow 16\% & \textuparrow 19\% & \textuparrow 17\% & \textuparrow 22\% & \textuparrow 12\% & \textuparrow 3\% & \textuparrow 11\%\\
Subsets (Uncertainty) & \textuparrow 19\% & \textuparrow 21\% & \textbf{\textuparrow 43\%} & \textuparrow 14\% & \textuparrow 25\% & \textuparrow 9\% & \textuparrow 2\% & \textuparrow 12\% \\
Weights (Uncertainty) & \textuparrow 11\% & \textuparrow 13\% & \textdownarrow 46\% & \textuparrow 7\% & \textuparrow 23\% & \textuparrow 5\% & \textuparrow 2\% & \textuparrow 7\%\\
\bottomrule
\end{tabular}
\label{table:fractures-classes}
\vspace{-1em}
\end{table*}
%%%%%%%%%%%%%%%%%%%%%%%%%%%%%%%%%%%%%%%%%%%%%%%%%%%%%%%%%%%%%%%%%%%%

%%%%%%%%%%%%%%%%%%%%%%%%%%%%%%%%%%%%%%%%%%%%%%%%%%%%%%%%%%%%%%%%%%%%%%
%% Table:Results (10 runs) MNIST IMBALANCE & noise
\begin{table*}[t]  % bht
\centering
\ra{1.5} 
\caption{Comparison of curriculum strategies driven by prior knowledge and uncertainty, under class-imbalance and label noise for the MNIST dataset. Mean error rate (\%). The highlighted values in bold correspond to the best strategy per scenario.}
\begin{tabular}{@{}l|c|cc|cc|cc@{}} \toprule
\multirow{2}{*}{} &
      \phantom{Baseline} &
      \multicolumn{2}{c|}{\textbf{Reorder}} &
      \multicolumn{2}{c|}{\textbf{Subsets}} &
      \multicolumn{2}{c}{\textbf{Weights}} \\
 & Baseline & Prior K. & Uncertainty & Prior K. & Uncertainty & Prior K. & Uncertainty \\
\midrule
Class-imbalance & 2.53 & 2.08 & 2.05 & \textbf{1.79} & 2.08 & 2.31 & 2.22 \\ 
\midrule
Label Noise & 9.46 & 8.76 & 8.42 & 8.28 & 7.24 & 8.49 & \textbf{5.42} \\
\bottomrule
\end{tabular}
\label{table:imbalance-noise}
%\vspace{-1em}
\end{table*}
%%%%%%%%%%%%%%%%%%%%%%%%%%%%%%%%%%%%%%%%%%%%%%%%%%%%%%%%%%%%%%%%%%%%%%

\section{Discussion} \label{sec:discussion}
In this work, we bring together several ideas from the literature and present them into a unified CL formulation. We experimentally demonstrate the effectiveness of ranking and scheduling training data for the challenging multi-class classification of proximal femur fractures. Most of the previous work \cite{Badgeley2019, Cheng2019,Wang2019WeaklySupervised} only target the fracture detection task, and Tanzi~\etal~\cite{tanzi2020hierarchical} does not obtain the same level of granularity. Our CL schemes achieve state-of-the-art results on the 7-class classification task. Furthermore, we also show the benefits of our CL strategies in a controlled set-up with MNIST dataset, specifically, under demanding scenarios such as class-imbalance, limited amounts of data and noisy annotations.

Inspired by classical CL, we leveraged prior knowledge to define the data scheduling elements. In our formulation, this prior knowledge only requires defining a scalar value per class. In case of multiple experts annotating the dataset, this knowledge can be derived from their intra- or inter-expert variability, or by asking the experts about the perceived difficulty of each class. One limitation of this approach is that the use of prior knowledge at the class level may be less informative for the CNN than at the sample level. More sophisticated sample-level strategies could be used. For example, images with noise or low resolution could be assigned a lower confidence score. When prior knowledge is not available, we have shown that uncertainty can be used to guide the optimization. We used MC dropout to estimate uncertainty. This has the advantage of not requiring any change in the CNN architecture, but it is computationally demanding. Indeed, the training time is doubled in this variant. Instead, one could investigate the use of a Dirichtlet distribution to parametrize the output of the network. Then, the behavior of such predictor could be interpreted from an evidential reasoning perspective, such as in subjective logic~\cite{Sensoy2018EvidentialDL, Jsang2018SubjectiveLogic}. Future research directions for defining the scoring function could be based on other uncertainty measures such as quantifying out-of-distribution samples~\cite{Hendrycks2016OOD} or evidence theory~\cite{Sensoy2018EvidentialDL}. We restricted our study to the predictive entropy of the model, which includes both aleatoric and epistemic uncertainty. We reckon that assessing separately each type of uncertainty could be advantageous for some applications. Moreover, if training time is not a concern, uncertainty does not only rank at the class but at the sample level. This scoring function is more appropriate for noisy annotations, since noise may affect individual samples and not entire classes. We restricted our experiments to individually applying each strategy, future work could try to combine them.

We evaluated three CL variants that consisted of reordering the whole training set, sampling subsets of data, or individually weighting training samples. Our CL schemes are compatible with any architecture and SGD training \cite{Weinshall2018CLbyTL}. They only require domain-specific knowledge or the estimated uncertainty for the definition of the scoring function, hence the curriculum. The reordering and subsets performances are very similar but if the dataset is too complex for the amount of available data (fractures), it seems better to keep the entire training set. We found similar performance when the curriculum probabilities were decayed towards a uniform distribution~\cite{Bengio2009CL} or maintained stable in our reorder and weights variants. Regarding the latter, we have proposed a simple and effective weighting scheme. In future work, we plan to explore other weighting strategies, \eg the focal loss~\cite{Lin2017FocalLoss}, which is well suited for class-imbalance scenarios, and the large margin loss
~\cite{Elsayed2018large}, which has been shown beneficial under limited amounts of data and when noisy labels are present. 

\section{Conclusions}
\label{sec:conclusions}
In this work, we have designed three CL strategies for the multi-class classification of proximal femur fractures. We validated the benefits of our approach reaching a performance comparable to state-of-the-art and experienced trauma surgeons. We have identified common scheduling elements in the literature and unified their formulation in our approach. We have proposed two types of ranking functions to prioritize training data, leveraging: prior knowledge and uncertainty. The best strategy for the classification of proximal femur fractures employed reordering with prior knowledge. In controlled experiments with the MNIST dataset, we have shown that the proposed method is effective for datasets with class-imbalance, limited or noisy annotations. From our experiments, we can conclude that for datasets of limited size or under the presence of class-imbalance, the use of the subsets variant can lead to an improved classification performance. One can either exploit prior knowledge to achieve a better performance, or if the computational cost is not an issue, leverage uncertainty. In the case of unreliable labels, we found that the more advantageous approach is the combination of weights with uncertainty. 

\bibliographystyle{IEEEtran}
\bibliography{IEEEabrv,refs}

\clearpage
\newpage

\section*{Supplementary Material}

%%%%%%%%%%%%%%%%%%%%%%%%%%%%%%%%%%%%%%%%%%%%%%%%%%%%%%%%%%%%%%%%%%%%%%
%% Table: p-values Fractures
\begin{table*}[]  % bht
\centering
\ra{1.5} %The higher the better. 
\caption{Statistical significance analysis for proximal femur fracture experiments. T-test with respect to baseline. P-values below 0.05 are bold-faced. }
\begin{tabular}{@{}c|cc|ccc|ccc@{}} \toprule
\multirow{2}{*}{\textbf{Prior knowledge}} &
      \multicolumn{2}{c|}{\textbf{Reorder}} &
      \multicolumn{3}{c|}{\textbf{Subsets}} &
      \multicolumn{3}{c}{\textbf{Weights}} \\
      & Anti-CL & CL & Random & Anti-CL & CL & Random & Anti-CL & CL \\
\midrule
7-class & \textbf{5.16E-04} & \textbf{3.03E-09} & 1.33E-01 & 5.61E-02	& \textbf{1.47E-06} & 3.26E-01 & 6.34E-01 & \textbf{1.48E-06} \\ 
3-class & \textbf{8.78E-04} & \textbf{5.43E-05} & 6.02E-01 & 7.20E-02 & \textbf{1.38E-02} & 4.73E-01 & \textbf{1.07E-02} & \textbf{1.57E-04} \\
\midrule \midrule
\multirow{2}{*}{\textbf{Uncertainty}} &
      \multicolumn{2}{c|}{\textbf{Reorder}} &
      \multicolumn{3}{c|}{\textbf{Subsets}} &
      \multicolumn{3}{c}{\textbf{Weights}} \\
      & Anti-CL & CL & Random & Anti-CL & CL & Random & Anti-CL & CL \\
\midrule
7-class & \textbf{7.10E-03} & \textbf{7.63E-05} & 1.33E-01 & \textbf{3.49E-03} & \textbf{2.27E-05} & 3.26E-01 & 4.11E-01 & \textbf{4.94E-05} \\ 
3-class & 4.80E-01 & \textbf{1.54E-02} & 6.02E-01 &	4.37E-01 & \textbf{2.98E-03} & 4.73E-01 & 3.10E-01 & \textbf{1.97E-02} \\
\bottomrule
\end{tabular}
\label{table:pvalues-fractures}
\vspace{-1em}
\end{table*}
%%%%%%%%%%%%%%%%%%%%%%%%%%%%%%%%%%%%%%%%%%%%%%%%%%%%%%%%%%%%%%%%%%%%%%

%%%%%%%%%%%%%%%%%%%%%%%%%%%%%%%%%%%%%%%%%%%%%%%%%%%%%%%%%%%%%%%%%%%%%%
%% Table: p-values MNIST
\begin{table*}[]  % bht
\centering
\ra{1.5} %The higher the better. 
\caption{Statistical significance analysis for MNIST experiments. T-test with respect to baseline. P-values are reported.}
\begin{tabular}{@{}c|cc|ccc|ccc@{}} \toprule
\multirow{2}{*}{\textbf{Prior knowledge}} &
      \multicolumn{2}{c|}{\textbf{Reorder}} &
      \multicolumn{3}{c|}{\textbf{Subsets}} &
      \multicolumn{3}{c}{\textbf{Weights}} \\
      & Anti-CL & CL & Random & Anti-CL & CL & Random & Anti-CL & CL \\
\midrule
30\% MNIST & 9.40E-01 & \textbf{8.70E-04} & \textbf{1.55E-04} & 1.62E-01 & \textbf{3.52E-06} & 3.29E-01 & \textbf{1.08E-04} & \textbf{6.03E-05} \\ 
50\%MNIST & \textbf{4.58E-04} & \textbf{8.37E-04} & 7.19E-02 & 8.59E-02 & \textbf{6.83E-05} & 2.04E-01 & 5.69E-02 & \textbf{7.23E-03} \\
100\%MNIST & \textbf{3.75E-03} & \textbf{1.22E-02} & 3.27E-01 & 3.88E-01 & \textbf{5.98E-04} & \textbf{3.09E-02} & 3.83E-01 & \textbf{9.95E-03} \\
\midrule \midrule
\multirow{2}{*}{\textbf{Uncertainty}} &
      \multicolumn{2}{c|}{\textbf{Reorder}} &
      \multicolumn{3}{c|}{\textbf{Subsets}} &
      \multicolumn{3}{c}{\textbf{Weights}} \\ & Anti-CL & CL & Random & Anti-CL & CL & Random & Anti-CL & CL \\
\midrule
30\% MNIST & 8.38E-01 & \textbf{7.16E-06} & 1.55E-04 & 7.77E-01 & \textbf{5.79E-07} & 3.29E-01 & 5.01E-01 & \textbf{3.11E-05} \\ 
50\% MNIST & 5.87E-01 & 3.77E-01 & 7.19E-02 & 1.15E-01 & \textbf{1.86E-04} & 2.04E-01 & 8.47E-02 & 6.43E-02 \\ 
100\% MNIST & \textbf{1.25E-02} & 6.41E-02 & 3.27E-01 & \textbf{3.44E-02} & \textbf{3.56E-03} & \textbf{3.09E-02} & 7.11E-01 & \textbf{2.49E-02} \\
\bottomrule
\end{tabular}x
\label{table:pvalues-mnist}
\vspace{-1em}
\end{table*}
%%%%%%%%%%%%%%%%%%%%%%%%%%%%%%%%%%%%%%%%%%%%%%%%%%%%%%%%%%%%%%%%%%%%%%

%%%%%%%%%%%%%%%%%%%%%%%%%%%%%%%%%%%%%%%%%%%%%%%%%%%%%%%%%%%%%%%%%%%%%%
%% Table: Fractures - subsets strategy - initial size
\begin{table*}[]  % bht
\centering
\ra{1.5} %The higher the better. 
\caption{$F_1$-score for the 7-class fracture classification, mean (median) and standard deviation for the subsets strategy with different initial subset sizes $N_S^{(0)}$. }
\begin{tabular}{@{}cc|cc@{}} \toprule
%\phantom{Baseline} & 
      \multicolumn{2}{c|}{\textbf{Prior knowledge-driven CL}} &
      \multicolumn{2}{c}{\textbf{Uncertainty-driven CL}} \\
      %Baseline & 
\midrule
      $N_S^{(0)}=25\%$ & $N_S^{(0)}=40\%$ & $N_S^{(0)}=25\%$ & $N_S^{(0)}=40\%$ \\
\midrule
%      56.32 (55.47) $\pm$ 3.19 &
      66.50 (66.02) $\pm$ 2.00 & 
      65.39 (65.76) $\pm$ 2.23 & 
      65.51 (66.32) $\pm$ 3.37 & 
      64.99 (65.63) $\pm$ 2.30 \\
\bottomrule
\end{tabular}
\label{table:subsets-init}
\vspace{-1em}
\end{table*}
%%%%%%%%%%%%%%%%%%%%%%%%%%%%%%%%%%%%%%%%%%%%%%%%%%%%%%%%%%%%%%%%%%%%%%

%%%%%%%%%%%%%%%%%%%%%%%%%%%%%%%%%%%%%%%%%%%%%%%%%%%%%%%%%%%%%%%%%%%%%%
%% Table: Fractures - subsets strategy - warm-up epochs Es
\begin{table*}[]  % bht
\centering
\ra{1.5} %The higher the better. 
\caption{$F_1$-score for the 7-class fracture classification, mean (median) and standard deviation for the subsets strategy with different number of epochs $E_S$ before considering the whole training set. }
\begin{tabular}{@{}ccc|ccc@{}} \toprule      \multicolumn{3}{c|}{\textbf{Prior knowledge-driven CL}} &
\multicolumn{3}{c}{\textbf{Uncertainty-driven CL}} \\
\midrule
      $E_S=5$ & $E_S=10$ &$E_S=20$ & $E_S=5$ & $E_S=10$ & $E_S=20$ \\
\midrule
63.68 (63.42) $\pm$ 3.15 & 
66.50 (66.02) $\pm$ 2.00 & 
66.09 (64.04) $\pm$ 1.24 & 
65.30 (65.78) $\pm$ 3.12 & 
65.51 (66.32) $\pm$ 3.37 & 
66.42 (66.68) $\pm$ 1.95 \\
\bottomrule
\end{tabular}
\label{table:subsets-epochs}
\vspace{-1em}
\end{table*}
%%%%%%%%%%%%%%%%%%%%%%%%%%%%%%%%%%%%%%%%%%%%%%%%%%%%%%%%%%%%%%%%%%%%%%

%%%%%%%%%%%%%%%%%%%%%%%%%%%%%%%%%%%%%%%%%%%%%%%%%%%%%%%%%%%%%%%%%%%%%%
%% Table: Fractures - weights strategy - batch sizes
\begin{table*}  % bht
\centering
\ra{1.5} %The higher the better. 
\caption{$F_1$-score for the 7-class fracture classification, mean (median) and standard deviation for the weights strategy with different batch sizes. }
\begin{tabular}{@{}ccc|ccc@{}} \toprule
%\phantom{Baseline} &
\multicolumn{3}{c|}{\textbf{Prior knowledge-driven CL}} &
\multicolumn{3}{c}{\textbf{Uncertainty-driven CL}} \\
\midrule
$B=16$ & $B=32$ & $B=64$ & $B=16$ & $B=32$ & $B=64$ \\
\midrule
65.35 (65.02) $\pm$ 2.97 & 
65.69 (65.95) $\pm$ 2.11 & 
64.65 (64.04) $\pm$ 1.56 & 
64.66 (65.76) $\pm$ 2.27 & 
66.92 (66.69) $\pm$ 2.18 & 
62.60 (62.96) $\pm$ 1.63 \\
\bottomrule
\end{tabular}
\label{table:weights-bsize}
\vspace{-1em}
\end{table*}
%%%%%%%%%%%%%%%%%%%%%%%%%%%%%%%%%%%%%%%%%%%%%%%%%%%%%%%%%%%%%%%%%%%%%%

%%%%%%%%%%%%%%%%%%%%%%%%%%%%%%%%%%%%%%%%%%%%%%%%%%%%%%%%%%%%%%%%%%%%%%
%%%%%%%%%%%%%%%%%%%%%%%%%%%%%%%%%%%%%%%%%%%%%%%%%%%%%%%%%%%%%%%%%%%%%

%%%%%%%%%%%%%%%%%%%%%%%%%%%%%%%%%%%%%%%%%%%%%%%%%%%%%%%%%%%%%%%%%%%%%%
\begin{figure}[ht]
    \centering
    \includegraphics[width=0.5\textwidth]{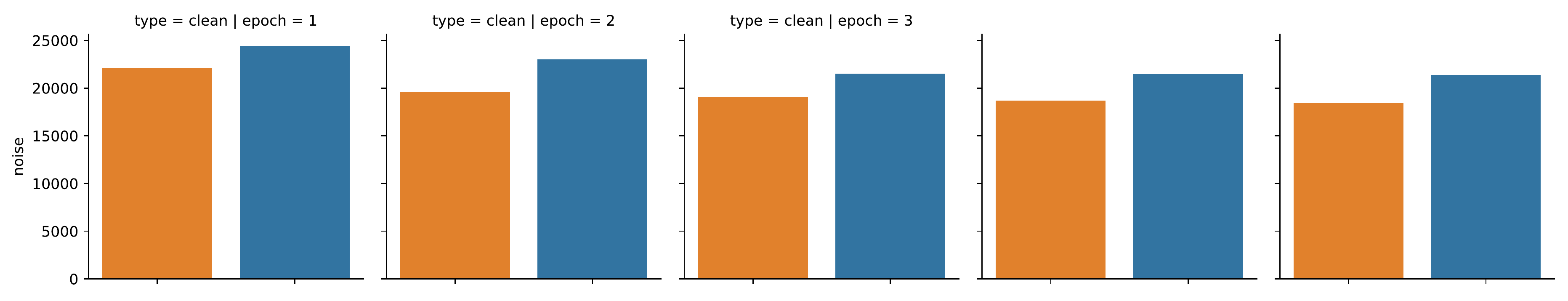}%
    \caption{Analysis of weights strategy under label corruption for MNIST dataset. Number of samples with a weight higher than the mean weight at that epoch. Random criterion and uncertainty are depicted in orange and blue, respectively.}
    \label{fig:mnist-noise}%
\end{figure}
%%%%%%%%%%%%%%%%%%%%%%%%%%%%%%%%%%%%%%%%%%%%%%%%%%%%%%%%%%%%%%%%%%%%%

\end{document}